\documentclass[10pt,twocolumn,letterpaper]{article}

\usepackage{cvpr}
\usepackage{times}
\usepackage{epsfig}
\usepackage{graphicx}
\usepackage{amsmath}
\usepackage{amssymb}

\usepackage[breaklinks=true,bookmarks=false]{hyperref}
\usepackage{cleveref}
\hypersetup{
  colorlinks, linkcolor=red
}

\cvprfinalcopy 

\def\cvprPaperID{2038} 

\begin{document}

\title{Optical Flow Requires Multiple Strategies (but only one network)}

\author{Tal Schuster, Lior Wolf and David Gadot\\
The School of Computer Science\\
Tel Aviv University\\
{\tt\small talschuster@gmail.com, wolf@cs.tau.ac.il, dedigadot@gmail.com}
 }

\maketitle

\begin{abstract}
   We show that the matching problem that underlies optical flow requires multiple strategies, depending on the amount of image motion  and other factors. We then study the implications of this observation on training a deep neural network for representing image patches in the context of descriptor based optical flow. We propose a metric learning method, which selects suitable negative samples based on the nature of the true match. This type of training produces a network that displays multiple strategies depending on the input and leads to state of the art results on the KITTI 2012 and KITTI 2015 optical flow benchmarks.
\end{abstract}

\section{Introduction}

In many AI challenges, including perception and planning, one specific problem requires multiple strategies. In the computer vision literature, this topic has gained little attention. Since a single model is typically trained, the conventional view is that of a unified, albeit complex, solution that captures all scenarios. Our work shows that careful consideration of the multifaceted nature of optical flow leads to a clear improvement in performing this task.

In optical flow, one can roughly separate between the small- and the large-displacement scenarios, and train model to apply different strategies to these different cases. The small displacement scenarios are characterized by relatively small appearance changes and require patch descriptors that can capture minute differences in appearance. The large displacement scenarios, on the other hand, require much more invariance in the matching process. 

\begin{figure}[t]
\centering
\begin{tabular}{cc}
(a) & \raisebox{-.5\height}{\includegraphics[width=0.80\linewidth]{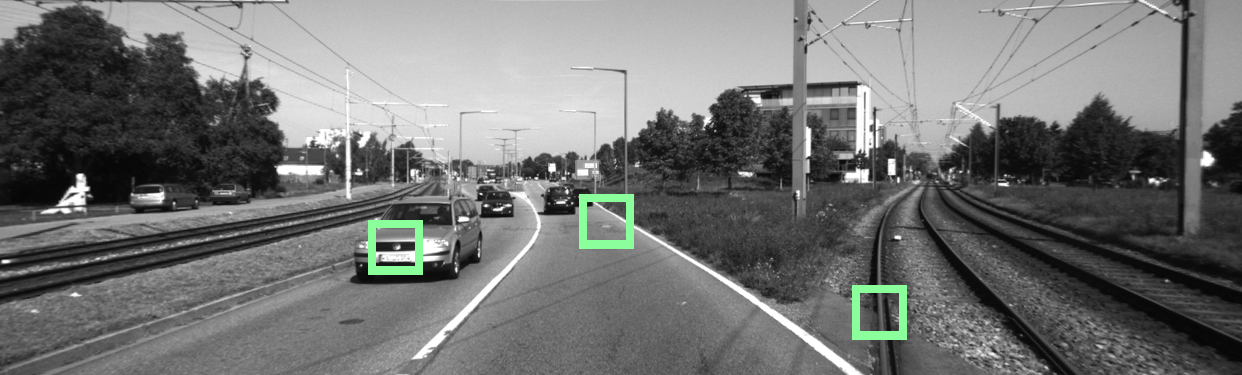}}\\
(b) & \raisebox{-.5\height}{\includegraphics[width=0.80\linewidth]{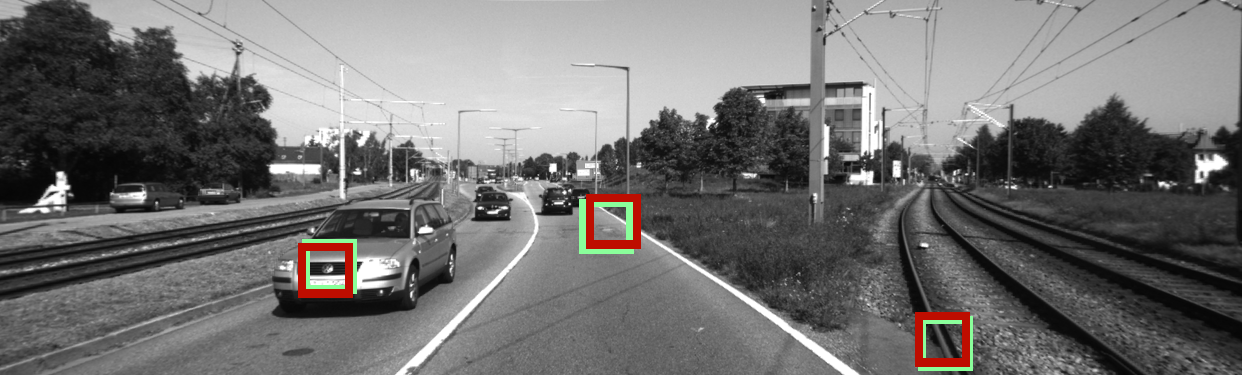}}\\
(c) & \raisebox{-.5\height}{\includegraphics[width=0.80\linewidth]{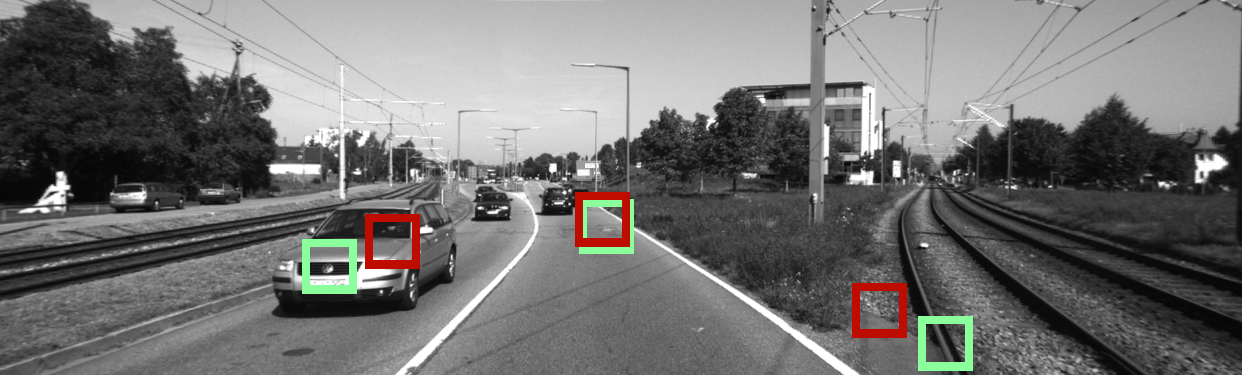}}\\
\\
\end{tabular}
\caption{Illustration of strategies for selecting negative samples.  (a) The first frame, in which some locations are marked. (b) In the baseline method~\cite{patchbatch}, negative samples are sampled close to the ground truth, regardless of the properties of the true match. Green are the true matches and Red are the negative samples.  (c) In the proposed method, the negative samples are sampled based on the displacement of the positive samples. Best viewed in color.}
\label{fig:one}
\end{figure}

State of the art methods in optical flow employ metric learning in order to learn the patch descriptors. We focus on the process of selecting negative samples during training and suggest two modifications. First, rather than selecting all negative samples close to the ground truth, we propose an \textit{interleaving learning} method that selects negative samples at a distance that match the amount of displacement that the true match (the positive sample) undergoes, as is illustrated in Fig.~\ref{fig:one}. Second, we suggest gradually increasing the difficulty of the negative samples during training. 

In the implementation of the second component, scheduling samples by difficulty, we combine two methods well known in the literature. The \textit{curriculum learning}  method~\cite{bengio2009curriculum} selects samples, stratified by difficulty, using a predefined order. The method of \textit{self-paced learning}~\cite{kumar2010self}  identifies a set of easy samples by their loss, and learns using only those samples.  The amount of samples defined as easy is increased over time. The \textit{Self-Paced-Curriculum-Interleaving} method we propose here combines in the selection process both the difficulty of a sample and its loss. However, in difference from the \textit{self-paced} method, no samples are excluded during training. Instead, we control the level of the difficulty of instances used for training by selecting negative samples of appropriate distances.

The pipeline employed for computing optical flow is similar to the PatchBatch method~\cite{patchbatch}. We slightly modify it by replacing the DrLIM loss with a Hinge loss.

\noindent Our main contributions in this work are:
\begin{itemize}
\itemsep0em
\item We analyze, for the first time, the need for multiple strategies in optical flow.
\item We propose a novel, psychologically inspired way to train a network to address multiple  scenarios at once.
\item We show how, in optical flow, our proposed new scheme translates to a simple, unexpected, heuristic.
\item We improve the PatchBatch\cite{patchbatch} pipeline itself.
\item State of the art results are demonstrated on the KITTI 2012 and KITTI 2015 benchmarks.
\end{itemize}

\section{Related work}



Many computer vision tasks require a pixel-wise image comparison (\eg image retrieval, object recognition, multi-view reconstruction). To allow for the comparison to be invariant to scale, rotation, illumination, \etc, image descriptors such as SIFT~\cite{lowe1999object}, SURF~\cite{bay2008surf}, HOG~\cite{dalal2005hog}, and~DAISY~\cite{tola2010daisy} have been used. Brox and Malik were the first to apply local descriptors to the problem of dense optical flow~\cite{brox2011large}. They found that the use of descriptors enables better performance for large displacement matching, but that the obtained solution has many outliers due to missing regularization constraints. In order to account for this, they used descriptors to build a sparse initial flow and interpolate it to a dense one using image smoothness assumptions. Following their success, many other models adopted the use of local descriptors~\cite{deepflow,discreteflow,TF+OFM,timofte2015sparse}.

With the advent of deep learning methods, CNNs were shown to be extremely powerful in the related problem of stereo matching~\cite{simo2015discriminative,zagoruyko2015learning}. For optical flow, a few CNN based models were proposed. In~\cite{walker2015static}, a CNN is used to predict the flow from a single static image. FlowNet~\cite{flownet} is the first end-to-end CNN for optical flow and showed competitive results. In the PatchBatch~\cite{patchbatch} pipeline, a CNN was used for extracting patch descriptors that are then used for matching via the PatchMatch~\cite{patchmatch} Nearest Neighbor Field (NNF) algorithms. It achieved state of the art performance in the KITTI benchmarks~\cite{kitti2012,kitti2015} as of last year.

While the use of descriptors has greatly improved overall performance and accuracy, methods keep failing with large displacements, as we further discuss in Section~\ref{sec:multi_faces}. To solve this problem, extensive efforts have been devoted to methods for the integration of descriptors with local assumptions~\cite{brox2011large,timofte2015sparse,discreteflow}. However, much less work was done in making the descriptors themselves more suitable for this scenario. A concurrent work~\cite{bailer2016cnn}, focused on decreasing the error for large displacements by down-sampling patches and adding a threshold to the loss function. However, this comes at the cost of reducing the accuracy obtained for small displacements.

In our work, we follow the PatchBatch pipeline and use a CNN to extract descriptors. We expand the work by analyzing different matching cases, specifically those of small and large displacements, and present a method for generating better matching descriptors for both cases.

\subsection{Learning for multiple strategies}

The need for multiple strategies was found in several vision problems where the basic trained model could not optimize the solution for all sub-categories. An example is the work of Antipov \etal~\cite{antipov_age} for age estimation. Unsatisfied by the accuracy of the model for children of age 0-12, they train a sub-model only for those ages and employ it to samples that are classified as this category by another model that is run first.

Another common case is in fine-grained classification, \eg determining the exact model of a car or a particular species of bird. The subtle differences between nearby species require, for example, to focus on specific body regions. However, different distinctions require different body parts and we can consider each body part as a separate decision strategy. 

In order to achieve the required accuracy, some methods perform object segmentation~\cite{krause2015fine} or part detection~\cite{krause2014learning} to limit the search of each sub-class to the most relevant body parts. A different approach was shown in~\cite{ge2016fine}, where several models were trained on different samples to create per class expert models. At test time, the answer with the highest confidence is chosen. The latter approach achieved better results due to each model leveraging all of the input data, and learning individually the required features to gain expertise in its task.

\subsection{Learning for varied difficulty levels}

\textit{Curriculum learning}~\cite{bengio2009curriculum}, inspired by the learning process of humans, was the first method to manipulate the order of samples shown to the model during training. Specifically, it is suggested to present the easy training samples first and the harder samples later, after performing stratification based on the difficulty level.

In \textit{self-paced learning}~\cite{kumar2010self}, instead of using a predefined order, the difficulty of each sample is dynamically estimated during training by inspecting the associated loss. On each epoch, only the easier samples are being learned from and their amount is increased with time until the entire data is considered. In the work of~\cite{spcl}, those two methods were combined to allow a prior knowledge of samples difficulty to be considered in the \textit{self-paced} iterations.

It was recently proposed to eliminate from the training process samples that are either too easy or too hard~\cite{boosting}. For this purpose, specific percentiles on the loss were employed. Samples which did not meet the loss criteria were put aside for a predefined number of epochs.

In the problem of optical flow, large displacements are known to be more challenging. Moreover, as we show in Section~\ref{sec:multi_faces}, the descriptor extraction strategy should differ by displacement. Due to the correlation between the difficulty level and the required strategy, applying the existing gradual learning methods could result in acquiring specific strategies in different training stages with the possibility of unwanted carryover. In Section~\ref{sec:learn_met}, we suggest novel learning techniques, which use all samples, support different strategies and apply an easy to hard order.

\section{The PatchBatch pipeline} \label{sec:pb}


\begin{figure}[t]
\centering
\includegraphics[width=0.8\linewidth]{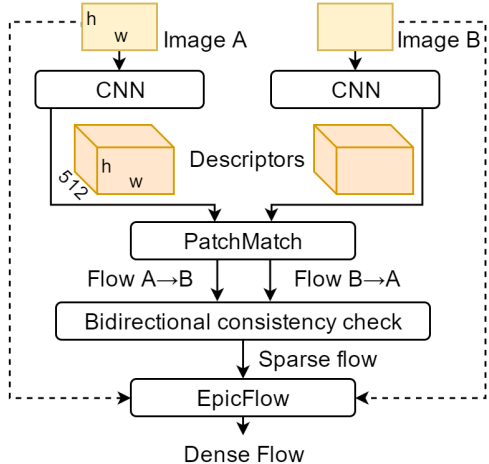}
\caption{Flow diagram of the PatchBatch pipeline. The same CNN is applied for patches from both images. PatchMatch~\cite{patchmatch} is applied twice in order to get both flow directions.}
\label{fig:pb_diagram}
\end{figure}

The PatchBatch (PB) pipeline, as described in Fig.~\ref{fig:pb_diagram}, consists of a CNN which generates per-pixel descriptors and an approximate nearest neighbor algorithm which is later used to compute the actual assignments. PatchBatch's ACCURATE network configuration generates descriptors with 512 float values. The assignment is computed by minimizing the $L_{2}$ distance between descriptor vectors. To create each pixel's descriptor, the CNN uses a patch as an input. In most of the CNN configurations described in PatchBatch, the input is a $51\times51$ patch centered around the examined pixel. The CNN uses the grayscale data of the patch to extract a descriptor as similar as possible to the one extracted for the matching pixel on the second image.

Using the generated descriptors, PatchMatch~\cite{patchmatch} (PM) algorithm is used to compute initial flow assignments. PM is applied in both flow directions and is followed by a bidirectional consistency check that allows elimination of non-consistent matches.

In the final step, the sparse-to-dense EpicFlow~\cite{epicflow} (EF) algorithm creates the final estimation using the sparse flow and the original raw images. We refer the reader to the PatchBatch~\cite{patchbatch} paper and the published code\footnotemark[1] for a more detailed description. 
\footnotetext[1]{\url{https://github.com/DediGadot/PatchBatch}}

\subsection{Architecture improvements} \label{sec:pb_modifications}

In this paper, we improve the CNN that generates the descriptors. We achieve this by several means. First, we adopt the suggestion, that was partially tested in the original PB paper~\cite{patchbatch}, to enlarge the patch size from $51\times51$ to $71\times71$ pixels. Second, to improve the training of the network we use two novelties: (1) We introduce a new learning method for multiple displacements detailed in Section~\ref{sec:learn_met}. (2) We modify the loss function and use a new form of the Hinge loss.
Third, we altered the initial random guess range of the PM algorithm on MPI-Sintel to be 100 instead of 10, to allow larger search distance and better utilization of our large displacements descriptors. For the KITTI benchmarks, this parameter remained unchanged (500). 

\begin{table*}
\begin{center}
\begin{tabular}{|l|c|c|c|c|c|c|c|c|c|}
\hline
Train set & 0-5 & 5-10 & 10-20 & 20-30 & 30-45 & 45-60 & 60-90 & 90-$\infty$ \\
\hline\hline
Baseline (all) & \textbf{2.32} & 7.32 & 5.32 & 9.38 & 25.21 & 50.43 & 67.32 & 216.39\\
\hline
$<$30 & 2.46 & \textbf{6.91} & \textbf{5.25} & \textbf{8.57} & 26.39 & 51.76 & 65.15 & 209.40\\
$>$30 & 3.03 & 9.07 & 5.64 & 10.29 & \textbf{24.74} & \textbf{46.81} & \textbf{56.69} & \textbf{199.61}\\
\hline
\end{tabular}
\end{center}
\caption{The increase of distractors with displacement and the success of models trained on a partial range, shown as average distractors amount by displacement range. The number of distractors for a given patch is the number of patches whose descriptors are within a smaller distance from it than the true match. Each column show the results for the Hinge+SD PB model trained on a specific displacement range.} \vspace*{-0.6\baselineskip}
\label{tab:distractors_by_set}
\end{table*}

\subsection{Hinge loss with SD}

Instead of the DrLIM~\cite{drlim} loss functions used in PatchBatch, we found the Hinge loss to achieve best results when integrated with our, further detailed, learning method. To allow the use of this loss, we construct the samples as triplets. For each patch, we collect a matching patch by the ground truth and a non-matching one. As a baseline, we use the same non-matching collecting method, which is a random patch up to 8 pixels from the matching one. 

We define the loss function as:
\begin{equation}
L_{H} = \frac{1}{n}\sum_{i=1}^{n} {max(0,m + D_{i,match} - D_{i,non-match})}
\end{equation} \label{eq:hinge}
where D is the $L_{2}$ distance of the examined patch descriptor from the matching or non-matching one.

In the PatchBatch paper, an addition of a standard deviation parameter was found to produce better distinction between matching and non-matching samples. With that inspiration, we apply a similar addition to the Hinge loss:
\begin{equation}
L_{H+SD} = \lambda L_{H} + (1 - \lambda)(\sigma_{D_{match}} + \sigma_{D_{non-match}})
\end{equation} \label{eq:hinge_sd}
We used $m=100$, $\lambda=0.8$ and a training set of $n=50k$ triplets for each epoch.

\section{Optical flow as a multifaceted problem} \label{sec:multi_faces}


It is clear by examining the results of the common optical flow benchmarks that optical flow methods are challenged by large displacements. In the MPI-Sintel~\cite{mpisintel}, where results are separated by the velocity of pixels, the current average end-point-error (EPE) of the top 10 ranked methods is 35.47 for velocities higher than 40, while it is about 1.01 for velocities lower than 10. In KITTI2015~\cite{kitti2015}, there is no published estimation by velocity. However, there is separation of foreground vs.\ background regions. The current average outliers percentage for the top 10 methods is 26.43\% for foreground versus 11.43\% for background, which, assuming foreground objects typically move faster than background, supports the same observation. When evaluating the baseline PatchBatch model on a validating set, we notice an error (percent of pixels with euclidean error $>$ 3) of 4.90\% for displacements smaller than 10 and 42.15\% for displacements larger than 40. 

\begin{figure}[t]
\centering
\includegraphics[width=1\linewidth]{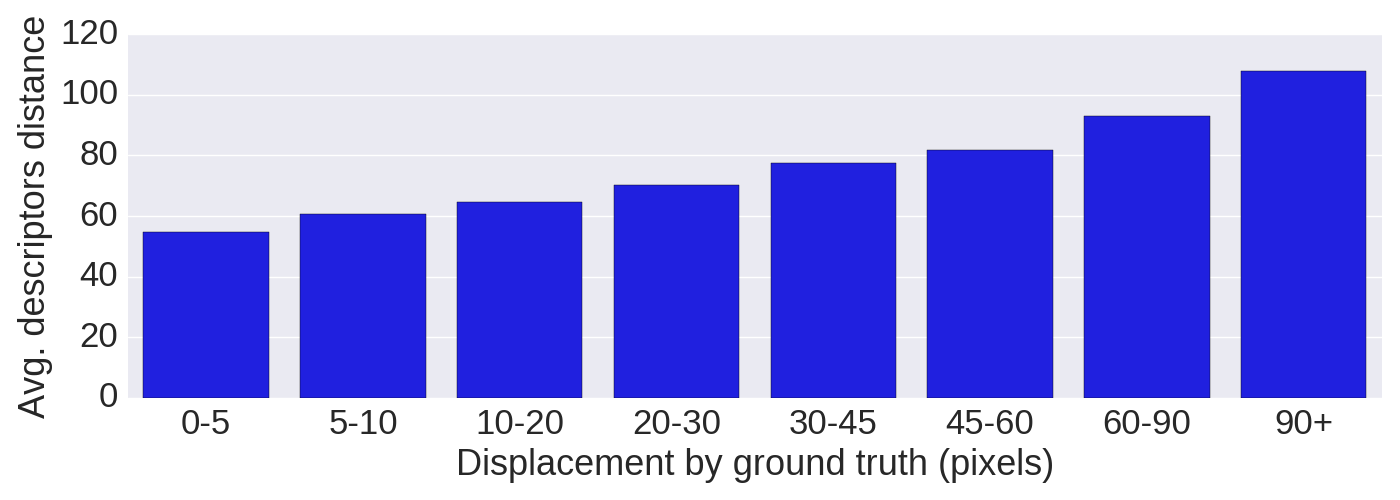}
\caption{Correlation of larger displacements with larger distances between the true matches. The average $L_{2}$ distance between descriptors of matching patches are shown grouped by displacement range. The descriptors were generated using a trained Hinge+SD PB model on the KITTI2012 benchmark.} \vspace*{-0.3\baselineskip}
\label{fig:l2_vs_disp}
\end{figure}

The challenge of matching at larger distances is exemplified in Fig.~\ref{fig:l2_vs_disp}, which shows the $L_2$ distance of the true match as a function of the ground truth displacement. Furthermore, as the distance increases, the average number of distractors in the second image, with higher similarity to patch in the first image than the true match, increases. This counting is performed in a radius of 25 pixels around the true match and is shown in Tab.~\ref{tab:distractors_by_set} under the Baseline training set. 



\subsection{Multiple strategies} \label{sec:need_strat}


When training the PatchBatch network only on displacements that are smaller than 30, we are able to improve most cases of small displacements, while, in most cases increasing the number of nearby distractors for large displacements. Conversely, training only on displacements larger than 30 pixels, achieved a lower amount of distractors for large displacements (Tab.~\ref{tab:distractors_by_set}). However, since there is no mechanism for selecting between the two networks, it is best to train one network that addresses both scenarios. Interestingly, when training just one network on all samples, the network seems to outperform the two specialized networks in the domain of very small displacements. This is probably a result of designing the PatchBatch method to excel in benchmarks that emphasize this category.



\begin{figure}[t]
\centering
\includegraphics[width=1\linewidth]{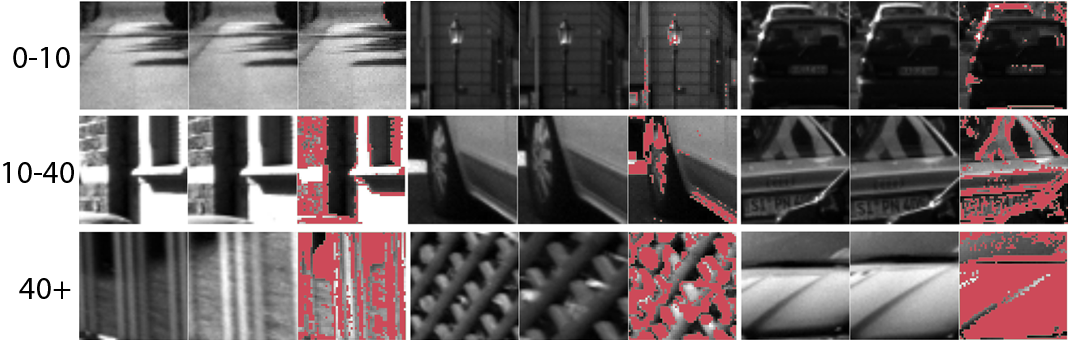}
\caption{Extent of pixel displacement is correlated with apparent differences in the KITTI benchmark. Samples are gathered in triplets, in which the matching pair is next to a display that depicts by red dots locations with $L_{1}$ distance larger than 0.2 between pixels value. Each row show examples from a displacement range that appears to the left. Best viewed in color.}  \vspace*{-0.6\baselineskip}
\label{fig:patch_samples}
\end{figure}

Large displacements are typically associated with larger differences in appearance, as  demonstrated in Fig.\ref{fig:patch_samples}.
Differences in the patch appearance for the small displacement case typically arise from objects moving within the patch faster than the middle pixel. In contrast, in large motions, we can expect much more pronounced changes in appearance due to the following: (1) As fast objects move, their background is more likely to change. (2) The view point changes more drastically, which leads to different object parts being occluded. (3) The distance and angle to light sources vary more quickly, leading to a change in illumination. (4) When a significant displacement occurs along the Z-axis of the camera, the object changes in both position and scale.

\section{Learning for multiple strategies and varying difficulty} \label{sec:learn_met}

As baseline methods, we apply gradual learning methods from the literature. For applying curriculum learning~\cite{bengio2009curriculum}, the samples need to be stratified by difficulty prior to training. Followed our previous findings, we define the difficulty level as the displacement value in the ground truth and increase the maximum displacement of the sample pool in each epoch which we call \textit{curriculum by displacement}.

Another curriculum implementation, which we call \textit{curriculum by distance}, would be to use samples with all displacement values for each epoch, and to start the training using false samples that have a large euclidean distance in the image from the true matching. Decreasing that distance with training should provide harder false samples with time.

We also implement a self-paced model by learning only from the easy samples in each epoch. Easiness here is measured per sample by requiring a loss that is lower than a threshold. The threshold increases over the training. 

\subsection{Interleaving learning}

We present a novel learning method for machine learning, motivated by the cognitive literature.\\
Both the curriculum learning approach as well as the self-paced one utilize the difficulty diversification of the samples and suggest to learn from easy to hard. While this idea might seem appealing, and does work in many machine learning problems, it could cause the network to become overly adapted to different aspects of the problem at different training stages. In optical flow, models must excel in the low displacement task in order to be competitive. Therefore, the shift of attention to harder and harder tasks is potentially detrimental. In addition, if different strategies are required, the carryover from the easy task to the more challenging ones is not obvious.

Our approach is motivated by psychological research. Kornell and Bjork, psychology researchers, found that for some cases, interleaving exemplars of different categories can enhance inductive learning~\cite{kornell2008learning}. Their tests showed that people learn better to distinguish classes, \eg bird species, by learning in an interleaving sample order rather than blocks of the same class. Another example would be sports training, in which it is common to interleave simple basic exercises with more complex ones, incorporating at least part of the complex movements from very early, and going back to the basic movements even after these are mastered.

The idiomatic way of training ML models is to randomize the feeding order of the samples. 
When perceptual strategies and difficulty levels are unrelated, the random process might be sufficient. However, when the samples that require some strategy A are consistently harder than the ones required for strategy B, the frequent loss related to the samples associated with A would mean that the strategy B would be deprived of a training signal.

To preserve a random order of strategies, and, at the same time, facilitate the penalty of harder samples, we suggest that the learning process should consider the difficulty of each sample. This could be done by either taking the difficulty of the sample into account while computing the penalty or, when training by pairs or triplets of samples, by controlling the composition of these small reference groups.

\subsection{Interleaving learning for optical flow}

\begin{figure}[t]
\centering
\includegraphics[width=0.8\linewidth]{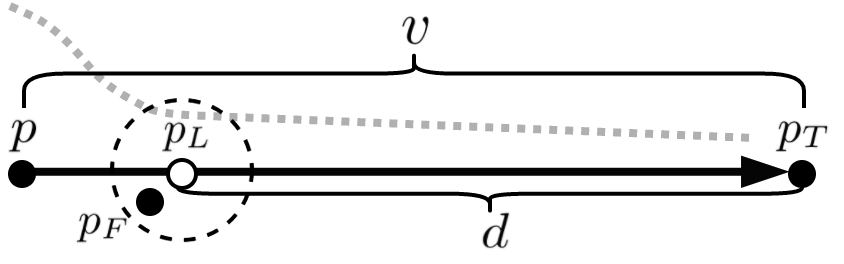}
\caption{Illustration of the false sample collecting methodology for interleaving learning. $p$ and $p_{T}$ represent a location in the first frame and its true matching from the second frame respectively. $p_{L}$ is sampled along the motion line ($p \rightarrow p_{T}$). The false sample ($p_{F}$) is randomly chosen from inside the dashed area that is 8 pixels from $p_{L}$. The dotted gray line represents the log-normal distribution from which $p_L$ is taken (mostly closer to $p$).}
\label{fig:sampling}
\end{figure}
The implementation of our method was done by using further patches as false samples for larger displacements. Thus, for the harder case of large displacements, we select false samples that should be easier to distinguish from the true ones and normalize the overall difficulty. From the strategy point of view, by presenting further away negatives for large displacements, the model learns to rely more on context and less on appearance changes for large displacements and conversely for small ones.

The chosen false sample distance is determined by:
\begin{equation}
d = v (1 - X) \qquad X\sim\log\mathcal{N(\mu,\sigma)}
\end{equation} \label{eq:sample_col}
\begin{equation}
P(X=x) = \dfrac{1}{\sigma x\sqrt{2\pi}}e^{(-\frac{(ln(x)-\mu)^{2}}{2\sigma^{2}})}
\end{equation} \label{eq:lognoraml}
where $v$ is the displacement of the matching pixels and $X$ is sampled from a log-normal distribution~\cite{lognormal}.

Using a log-normal distribution, allows us to take samples mostly relative to the exemplar motion while also providing a small amount of harder samples. We used $\mu=0$ and $\sigma=1$ as parameters and after sampling values for all of the batch samples, they were normalized to $[0,1]$.

To implement this method in our learning process, we collect the false sample along the line connecting the original and the destined coordinates of the patch. Specifically, we randomly select a sample from a radius of up to 8 pixels from the point with distance $d$ from the true match on that line, in the direction of the position in the first image (see Fig.~\ref{fig:sampling}). Interestingly, for the purpose of creating dual strategy descriptors, it does not matter whether the samples are from along the motion line. However, in our experiments, it turned out that sampling this way slightly helps the subsequent PM step. This is probably because PM initially searches in a random distance from the original patch position. By taking a false match that is closer to the original location, we help eliminate those samples.

\begin{table*}[t]
\begin{center}
\begin{tabular}{|l|c|c|c|c|c|c|c|c|c|c|c|}
\hline
\multicolumn{1}{|c|}{Model / Learning} & \multicolumn{2}{c|}{Error percent} & \multicolumn{9}{c|}{Distractors amount by displacement range} \\
\multicolumn{1}{|c|}{\qquad \quad \ method} & \multicolumn{1}{c}{post PM} & \multicolumn{1}{c|}{post EF} & \multicolumn{1}{c}{0-5} & \multicolumn{1}{c}{5-10} & \multicolumn{1}{c}{10-20} & \multicolumn{1}{c}{20-30} & \multicolumn{1}{c}{30-45} & \multicolumn{1}{c}{45-60} & \multicolumn{1}{c}{60-90} & \multicolumn{1}{c}{90-$\infty$} & \multicolumn{1}{c|}{All} \\
\hline\hline
CENT ~\cite{patchbatch} & 9.93\% & 5.19\% & 3.31 & 15.34 & 16.87 & 27.61 & 48.28 & 69.19 & 92.62 & 209.13 & 32.86\\
CENT+SD~\cite{patchbatch} & 8.91\% & 4.85\% & 4.33 & 16.7 & 12.29 & 19.92 & 38.20 & 60.69 & 81.22 & 216.02 & 28.67\\
CENT+SD / \textbf{Inter} & 8.75\% & 4.70\% & 2.61 & 10.50 & 8.64 & 15.29 & 30.38 & 42.87 & 66.16 & 137.81 & 20.73\\
\hline
Hinge & 7.78\% & 5.18\% & 1.93 & 8.14 & 5.81 & 10.98 & 31.95 & 50.97 & 73.24 & 185.81 & 21.40\\
Hinge+SD & 7.74\% & 4.85\% & 2.32 & 7.32 & 5.32 & 9.38 & 25.21 & 50.43 & 67.32 & 216.39 & 20.51\\
Hinge+SD / Neg-mining & 7.53\% & 5.00\% & 3.06 & 6.19 & 5.41 & 10.52 & 26.88 & 51.33 & 70.29 & 210.34 & 20.96 \\
Hinge+SD / Cur.\ by disp & 7.67\% & 4.83\% & 2.71 & 8.61 & 5.26 & 10.26 & 14.76 & 48.88 & 65.15 & 220.13 & 20.67 \\
Hinge+SD / Cur.\ by dist & 7.47\% & 4.93\% & 2.83 & 8.66 & 5.25 & 10.35 & 23.62 & 45.82 & 63.69 & 197.82 & 19.70 \\
Hinge+SD / Self-Paced & 8.75\% & 5.23\% & 2.88 & 9.35 & 6.84 & 13.74 & 34.09 & 57.46 & 80.8 & 198.97 & 23.93\\
Hinge+SD / Anti-Inter & 14.53\% & 8.30\% & 2.98 & 9.12 & 13.36 & 20.63 & 37.69 & 42.41 & 81.41 & 132.03 & 24.11\\
Hinge+SD / \textbf{Inter} & 6.60\% & 4.41\% & 1.41 & 5.57 & 3.07 & 6.31 & 15.6 & 28.52 & 43.46 & 127.65 & 12.61\\
Hinge+SD / \textbf{SPCI} & 6.64\% & 4.37\% & 1.40 & 5.04 & 3.46 & 6.56 & 15.11 & 27.13 & 42.72 & 130.17 & 12.50\\
\hline
Hinge+SD+PS71 & 7.34\% & 4.76\% & 1.96 & 5.44 & 5.28 & 11.8 & 22.76 & 42.3 & 67.27 & 190.3 & 18.91\\
Hinge+SD+PS71 / \textbf{Inter} & 6.17\% & 4.35\% &\textbf{ 1.00 }& 3.96 & 2.22 & 4.11 & 11.33 & \textbf{20.87} & 32.53 & \textbf{119.74} & 9.80\\
Hinge+SD+PS71 / \textbf{SPCI} & \textbf{6.12}\% & \textbf{4.27}\% & 1.02 & \textbf{3.42} & \textbf{2.16} & \textbf{3.52} & \textbf{10.55} & 21.28 & \textbf{32.17} & 119.98 & \textbf{9.54}\\
\hline
\end{tabular}
\end{center} \vspace*{-0.4\baselineskip}
\caption{Architecture and learning method comparison by the output error of the PatchMatch and EpicFlow steps in the pipeline and by distractors amount. SD symbols the addition of the standard deviation to the loss function, PS71 is for using a patch size of $71\times71$ pixels. Neg-mining was implemented as described in~\cite{simo2015discriminative} with a factor of 2. See Section~\ref{sec:learn_met} for an explanation of the other learning methods. The error is the percent of pixels in the validation set with euclidean error $>$ 3 pixels. Distractors are calculated as described in Section~\ref{sec:multi_faces}.} \vspace*{-0.8\baselineskip}
\label{tab:val_results}
\end{table*} 

\subsection{Self-Paced Curriculum Interleaving learning}

Given the interleaving learning method, which, unlike curriculum learning employs all samples at once, we can expand it by adding a dynamic control on the difficulty level. In order to maintain the category diversity, we simply modify the distance equation for epoch $i$ to:
\begin{equation} 
d_{i} = v (1 - X - R_{i})
\end{equation}  
where $R_{i}$ is define as:
\begin{equation} \label{eq:spci}
R_{i} = \underbrace{\frac{i}{m}}_\textrm{curriculum} \cdot  \underbrace{\max(0,1-\frac{l_{i-1}}{l_{init}})}_\textrm{self-paced}
\end{equation}
and $m$ is the total epoch amount, $l_{i}$ is the validation loss on epoch $i$ and $l_{init}$ is some initial loss to compare. We defined $l_{init}$ as the loss on epoch number 5. Until that epoch, self-pacing is not applied.

The curriculum addition enhances the global difficulty of false samples in each iteration by shorting the taken distance and, therefore, integrates an instructor-driven approach assuming the student will handle more difficult tasks with time. To add a student-driven portion, we use the self-paced component which allows a feedback from the model to influence the difficulty of the next iteration. Integrating all of this together, we get a learning method that learns all strategies simultaneously and in which the difficulty is increased over iterations and with a success feedback.

\section{Experiments} \label{sec:eval}

We perform two families of experiments. First, MNIST recognition experiments are presented as a testbed for the learning schemes. Then, the main set of experiments is performed on the specific problem of optical flow.

\subsection{MNIST}
In order to validate our learning methods on a task different from optical flow, we used the MNIST handwritten digit database~\cite{mnist}. This data set consists of images showing a digit from 0 to 9 with their true label. We divided the data into two different classes -- class L contains digits $0..4$ and class H contains $5..9$ . To enable difficulty differentiation between samples, random noise was added to the top half of the images of $H$ and to the bottom part of the $L$ images. Furthermore, images from class $H$ were rotated by a random angle of $[0,45]$ degrees with correlation to the noise amount, such that, samples that are more noisy are also rotated in larger angles.

While referring noisier samples as harder, we trained a model using several methods. As curriculum learning, harder samples were added to the training pool in each epoch. In the self-paced model, the hardness of the samples to learn from was derived from the loss. Interleaving was implemented by using all of the noise range level in each epoch with a fewer noised samples for the harder $H$ class against more for $L$ class. An integration of interleaving with Curriculum and Self-Paced methods was also used by increasing the the amount of the noised $H$ samples in each epoch. As can be seen in Tab.~\ref{tab:mnist}, interleaving produced the greatest improvement and SPCI attained the best results. 

\begin{table}[t]
\begin{center}
\begin{tabular}{|l|c|c|}
\hline
Method & $L$ & $H$\\
\hline\hline
Random order & 97.98\% & 82.24\%\\
Curriculum & 98.10\% & 87.89\%\\
Self-Paced & 98.26\% & 88.33\%\\
\textbf{Interleaving} & 98.26\% & 95.00\%\\
\textbf{Interleaving}+Curriculum & 98.30\% & 95.62\%\\
\textbf{Interleaving}+SP & 98.14\% & 95.31\%\\
\textbf{SPCI} & \textbf{98.38}\% & \textbf{96.33}\%\\
\hline
\end{tabular}
\end{center}
\caption{The improvement of results on the MNIST experiment using interleaving methods. Column $L$ shows the results on digits $[0,4]$  with random noise on the image bottom, and column $H$ shows the results on digits $[5,9]$ rotated randomly by  $0$ to $45$ degrees with random noise at the top of the image .} \vspace*{-0.4\baselineskip}
\label{tab:mnist}
\end{table}

\subsection{Optical flow}
To evaluate our work, we use the three most competitive optical flow benchmarks - KITTI2012~\cite{kitti2012}, KITTI2015~\cite{kitti2015} and MPI-Sintel~\cite{mpisintel}. We use their data to conduct a series of experiments to measure the effect of each of our contributions and to submit our best results to compare with other methods.

By training the different models on a subset of 80\% from the KITTI2012 dataset for 500 epochs and testing the results on the remaining 20\% image pairs, we show a comparison of the models summarized in Tab.~\ref{tab:val_results}. Note that lower PatchMatch (PM) error is not always correlated with lower EpicFlow (EF) error because of the bidirectional consistency check that excludes some inconsistent results to generate a sparse flow as an input for EF.

Observing Tab.~\ref{tab:val_results}, one can notice that the use of the Hinge loss instead of CENT~\cite{patchbatch}, improved the PM results and has no such effect on the final EF output. However, combining with the batch standard deviation term (SD) and our interleaving learning (Inter) leads to an advantage of the Hinge loss. Our interleaving learning method outperforms both Curriculum learning and Self-Paced learning. The SPCI technique contributes an additional improvement. Integrating all of our architecture modifications with SPCI produces the lowest error percent on the validation set with a major improvement on the initial baseline. Moreover, the amount of nearby distractors with descriptors that are more similar to the original patch than the true match is reduced to one third of the baseline.  

As a sanity-check experiment we evaluate an \textit{Anti-Interleaving} method. 
In this method, negative matches from different ranges were also used. However, the ratio was inverted -- true matches of small displacements were matched with false samples with large distances and vice versa. The high error of this model, as can be seen in Tab.~\ref{tab:val_results}, implies that the use of different ranges for false matches was not the main benefit of the interleaving method and it is the correlation with the displacement values that is the crucial factor.

\subsubsection{Sensitivity to appearance change}

\begin{table}[t]
\begin{center}
\begin{tabular}{|l|c|c|c|}
\hline
Method & 5 - 10 & 10 - 40 & 40 - $\infty$ \\
\hline\hline
Baseline &  95.01\% & 97.61\% & 97.83\%\\
Cur.\ by displacement* & 96.82\% & 98.56\% & 101.04\%\\
Cur.\ by distance* & 98.40\% & 98.32\% & 100.29\%\\
Self-Paced* & 93.66\% & 93.67\% & 99.78\%\\
Anti-Interleaving & 105.29\%& 116.34\% & 103.26\%\\
\textbf{Interleaving} & 97.32\% & 94.67\% & 93.71\%\\
\textbf{Interleaving}+Cur.** & 96.40\% & 95.39\% & 95.24\%\\
\textbf{Interleaving}+SP** & 95.82\% & 95.38\% & 93.61\%\\
\textbf{SPCI} & 96.02\% & 92.66\% & 90.11\%\\
\hline
\end{tabular}
\end{center}
\caption{Learning method comparison by descriptor sensitivity to location movement for different displacement ranges, measured by dividing the average distance of the descriptors of 5 pixels neighbor patches associated with a certain displacement range with the average obtained at for displacements smaller than 5 pixels. Methods marked with * were implemented as described in the beginning of Section~\ref{sec:learn_met} and the ones marked with ** were trained like SPCI, but applying only one multiplier in Eq.~\ref{eq:spci}. Using only gradual methods seems not to have any tendency relating to displacement value. In contrast, the interleaving models have learned to progressively decrease sensitivity for larger values.}
\label{tab:movement}
\end{table}

Part of what the networks learn is to behave differently to patches with different {\em expected} displacements. Those patches that are similar to patches that are associated with small displacements are treated differently than those which were associated, in the training set, with large displacements. To illustrate this, and compare the various learning methods, we explore the model behavior on nearby patches from the {\em same image} for varied displacement ranges. First, we measure the average distance $\bar d_{0-5}$ of a patch descriptor from that of a patch that is 5 pixels away for pixels which undergo a displacement of up to 5 pixels. Note that for a $51 \times 51$ patch, only 18\% of the pixels were completely replaced in such a small displacement. Then, we repeat this to patches from various displacement ranges, taking again the average distance from a patch of 5 pixels away. To normalize, we divide this average distance by the first average $\frac{\bar d_{L-H}}{\bar d_{0-5}}$, for $(L,H) \in \{(5,10),(10,40),(40,\inf)\}$.

The results in Tab.~\ref{tab:movement} show that while the PatchBatch original model reacts almost similarly for all displacement ranges, interleaving trained models have learned to be less sensitive to appearance changes for larger displacements. Moreover, using only gradual learning, leads to high sensitivity across all ranges. This can be the result of the carry-on from the early learning stages on small displacements where appearance sensitivity is more valuable. 

\subsubsection{Benchmarks results} \label{sec:bench_results}
We train our model on three datasets and submit the results of each benchmark on the respectively trained model. Our results are directly comparable with the PatchBatch model, since we use the same procedure as theirs -- Training the CNN for 4000 epochs on 80\% of the training set and choosing the best configuration by selecting the one with the lowest validation error on samples from the remaining 20\% of the data. 

The results can be seen in Tab.~\ref{tab:kitti12_results},~\ref{tab:kitti15_results},~\ref{tab:mpi_results}. We succeed in improving results in all three benchmarks and achieve state of the art results for KITTI2012~\cite{kitti2012} and KITTI2015~\cite{kitti2015}. 

We evaluate our method only against methods not using additional information for the flow estimation, including those methods which used semantic segmentation.

On KITTI2015, as can be seen on Tab.~\ref{tab:kitti15_results}, we reduced the error of both foreground and background areas, obtaining the lowest error for both cases. The increased accuracy for both regions is correlated with our previous experiments and corroborate our claim of extracting better descriptors for all scenarios.


\begin{table}[t]
\begin{center}
\begin{tabular}{|l|c|}
\hline
Method & Out-Noc\\
\hline\hline
{\bf Imp.\ PatchBatch+SPCI} &  \textbf{4.65}\% \\
CNN-HPM~\cite{bailer2016cnn} & 4.89\% \\
{\bf Imp.\ PatchBatch} &  4.92\% \\
PatchBatch+PS71~\cite{patchbatch} &  5.29\% \\
PatchBatch~\cite{patchbatch} &  5.44\%\\
PH-Flow~\cite{phflow} & 5.76\%\\
FlowFields~\cite{flowfields} & 5.77\%\\
CPM-Flow~\cite{huefficient} & 5.79\% \\
\hline
\end{tabular}
\end{center}
\caption{Top 8 published KITTI2012 Pure Optical Flow methods as of the submission date. Imp.\ PatchBatch denotes the PB pipeline with the improvements described in Section~\ref{sec:pb}. Out-Noc is the percentage of pixels with euclidean error $>$ 3 pixels out of the non-occluded pixels.} \vspace*{-0.6\baselineskip}
\label{tab:kitti12_results}
\end{table} 

\begin{table}[t]
\begin{center}
\begin{tabular}{|l|c|c|c|}
\hline
Method & Fl-bg & Fl-fg & Fl-all\\
\hline\hline
{\bf Imp.\ PatchBatch+SPCI} & \textbf{17.25}\% & \textbf{24.52}\% & \textbf{18.46}\%\\
CNN-HPM~\cite{bailer2016cnn} & 18.90\% & 24.96\% & 19.44\%\\
PatchBatch~\cite{patchbatch} & 19.98\% & 30.24\% & 21.69\%\\
DiscreteFlow~\cite{discreteflow} & 21.53\% & 26.68\% & 22.38\%\\
CPM-Flow~\cite{huefficient} & 22.32\% & 27.79\% & 23.23\%\\
FullFlow~\cite{chen2016full} & 23.09\% & 30.11\% & 24.26\%\\
EpicFlow~\cite{epicflow} & 25.81\% & 33.56\% & 27.10\%\\
DeepFlow~\cite{deepflow} & 27.96\% & 35.28\% & 29.18\%\\

\hline
\end{tabular}
\end{center}
\caption{Top 8 published KITTI2015 Pure Optical Flow methods as of the submission date. Imp.\ PatchBatch denotes the PB pipeline with the improvements described in Section~\ref{sec:pb}. Fl-all is the percentage of outliers (pixels with euclidean error $>$ 3 pixels). Fl-bg, Fl-fg are the percentage of outliers only over background and foreground regions respectively.} 
\label{tab:kitti15_results}  
\end{table}

\begin{table}[t]\vspace*{-0.5\baselineskip}
\begin{center}
\begin{tabular}{|l|c|c|c|c|c|}
\hline
Method & EPE & Fl & s0-10 & s40+\\
\hline\hline
FlowFields+~\cite{flowfields} & \textbf{5.71}  & 8.14\% & 1.31 & \textbf{34.17}\\
DeepDiscreteFlow~\cite{deepDiscreteFlow} & 5.73 & \textbf{7.30}\% & 0.96 & 35.82\\
SPM-BPv2~\cite{spmbp} & 5.81 & 9.17\% & 1.05 & 35.12\\
FullFlow~\cite{chen2016full} & 5.90 & 9.55\% & 1.14 & 35.59\\
CPM-Flow~\cite{huefficient} & 5.96 & 8.31\% & 1.15 & 35.14\\
GlobalPatchCollider~\cite{Wang_2016_CVPR} & 6.04 & 10.21\% & 1.10 & 36.45\\
DiscreteFlow~\cite{discreteflow} & 6.08 & 9.52\% & 1.07 & 36.34\\
{\bf Imp.\ PatchBatch+Inter} & 6.22 & 8.11\% & 0.91 & 39.91\\
{\bf Imp.\ PatchBatch+SPCI} & 6.24 & 7.89\% & 0.88 & 40.07\\
EpicFlow~\cite{epicflow} & 6.28 & 11.26\% & 1.13 & 38.02\\
FGI~\cite{li2016fast} & 6.61 & 12.34\% & 1.15 & 39.98\\
TF+OFM~\cite{TF+OFM} & 6.73 & 11.35\% & 1.51 & 39.76\\
Deep+R~\cite{deep+r} & 6.77 & 13.71\% & 1.16 & 41.69\\
PatchBatch  ~\cite{patchbatch} & 6.78 & 8.66\% & \textbf{0.72} & 45.86\\
\hline
\end{tabular}
\end{center}
\caption{Comparison of our models with the top methods for the MPI-Sintel benchmark as of the submission date. Imp.\ PatchBatch denotes the PB pipeline with the improvements described in Section~\ref{sec:pb}. The EPE (end-point-error) is averaged over all the pixels and the two right columns contain only the EPE of pixels within the displacement range mentioned in the title. The Fl column presents an evaluation of the the outlier percentage, which, although not provided by this benchmark, was calculated from the error figures presented for each scene that have higher pixel values for larger errors. Fl is the percentage of pixels with a value larger than 120.} \vspace*{-0.5\baselineskip}
\label{tab:mpi_results}
\end{table}
In contrast to the error percent measurement of the KITTI benchmarks, MPI-Sintel uses an end-point-error (EPE) one. Compared to the original PatchBatch model, (Tab.~\ref{tab:mpi_results}) we succeed in preserving a low EPE for small displacements while significantly reducing it for large ones. Our model does not  achieve the best results when using the EPE measurement. However, when considering the percentage of large error displacements, as calculated from the error images, our SPCI model is second best and our interleaving model is third.

Our trained models are available on the PatchBatch GitHub repository.

\section{Conclusions}

Common sense dictates that most of the perceptual tasks are heterogeneous and require multiple strategies. The literature methods address training in accordance with the difficulty of specific samples. In our work, we show, for the first time, how to address both multiple sub-tasks and varying difficulty. The two are not independent -- some sub-tasks are harder than others, and our interleaving methods address this challenge.

Using the proposed novel methods, we are able to improve a recently proposed optical flow model and obtain state of the art results on the two most competitive real-world benchmarks.

\section*{Acknowledgments}
This research is supported by the Intel Collaborative Research Institute for Computational Intelligence (ICRI-CI).

{\small
\bibliographystyle{ieee}
\bibliography{flow}
}

\end{document}